\documentclass[letterpaper, 10 pt, conference]{ieeeconf}
\IEEEoverridecommandlockouts
\usepackage{cite}
\usepackage{amsmath,amssymb,amsfonts}
\usepackage{algorithmic}
\usepackage{graphicx}
\usepackage{textcomp}
\usepackage{graphicx}
\usepackage[table]{xcolor}
\usepackage{xcolor}
\usepackage[hidelinks]{hyperref}
 \usepackage{multirow,multicol, makecell, booktabs}
\usepackage[export]{adjustbox}
\usepackage{subcaption}
\usepackage{xurl}
\usepackage{makecell}
\usepackage{array}

\newcolumntype{L}[1]{>{\raggedright\let\newline\\\arraybackslash\hspace{0pt}}m{#1}}
\newcolumntype{C}[1]{>{\centering\let\newline\\\arraybackslash\hspace{0pt}}m{#1}}
\newcolumntype{R}[1]{>{\raggedleft\let\newline\\\arraybackslash\hspace{0pt}}m{#1}}

\def\BibTeX{{\rm B\kern-.05em{\sc i\kern-.025em b}\kern-.08em
    T\kern-.1667em\lower.7ex\hbox{E}\kern-.125emX}}
\begin{document}

\title{Evaluation of Position and Velocity Based Forward Dynamics Compliance Control (FDCC) for Robotic Interactions in Position Controlled Robots}

\author{Mohatashem Reyaz Makhdoomi\textsuperscript{1}, Vivek Muralidharan\textsuperscript{1},\\ Juan Sandoval\textsuperscript{2}, Miguel Olivares-Mendez\textsuperscript{1} and Carol Martinez\textsuperscript{1}}
\maketitle

\footnotetext[1]{Space Robotics (SpaceR) Research Group, Interdisciplinary Research Centre for Security, Reliability and Trust (SnT), University of Luxembourg, 1855, Luxembourg 
        \{{\tt mohatashem.makhdoomi},
        {\tt vivek.muralidharan},
        {\tt miguel.olivaresmendez},
        {\tt carol.martinezluna\}@uni.lu}%
}
\footnotetext[2]{
Department of GMSC, PPRIME Institute, CNRS, ENSMA, University of Poitiers, 86073 Poitiers, France

         {\tt juan.sandoval@univ-poitiers.fr}%
}

\maketitle

\begin{abstract}
In robotic manipulation, end-effector compliance is an essential precondition for performing contact-rich tasks, such as machining, assembly, and human-robot interaction. Most robotic arms are position-controlled stiff systems at a hardware level. Thus, adding compliance becomes essential. Compliance in those systems has been recently achieved using Forward dynamics compliance control (FDCC), which, owing to its virtual forward dynamics model, can be implemented on both position and velocity-controlled robots. This paper evaluates the choice of control interface (and hence the control domain), which, although considered trivial, is essential due to differences in their characteristics. In some cases, the choice is restricted to the available hardware interface. However, given the option to choose, the velocity-based control interface makes a better candidate for compliance control because of smoother compliant behaviour, reduced interaction forces, and work done. To prove these points, in this paper FDCC is evaluated on the UR10e six-DOF manipulator with velocity and position control modes. The evaluation is based on force-control benchmarking metrics using 3D-printed artefacts. Real experiments favour the choice of velocity control over position control.

\end{abstract}

\section{Introduction}

\begin{figure}[h!]
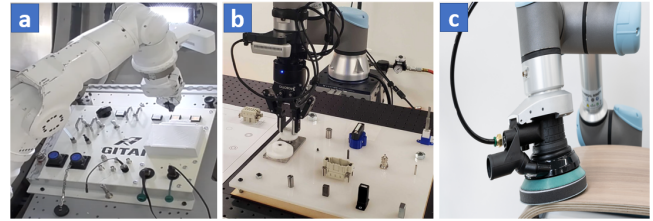

\setlength{\belowcaptionskip}{-15pt}
\centering
       \adjincludegraphics[width=0.99\linewidth,trim={0cm 0cm 0.0cm 0cm},clip]{Figures/scenarios.png} 
    \caption{End-effector Compliance in Terrestrial and Space Applications. (a) ISS task-board assembly \cite{gitai_space_taskboard} (b) NIST Industrial Taskboard \cite{nist_taskboard} (c) Surface-Finishing \cite{surface_finshing}}
    \label{fig:scenario}
\end{figure}

The capability of a robotic arm to handle interactions with the environment is one of the most fundamental requirements for successful robotic manipulation\cite{siciliano2009force}.   
Applications such as inspection; contact operations such as pushing buttons, twisting knobs and dials; instrumentation and parts assembly (insertion and removal of connectors, screwing/unscrewing operations), and human-robot interaction, nowadays common to both space and terrestrial domains (see Fig. \ref{fig:scenario}), require compliance capabilities to deal with high-precision and unstructured and uncertain environments \cite{papadopoulos2021robotic, buckmaster2008compliant}.

The interaction state is measured by the contact force acting at the manipulator's end-effector, provided it is equipped with a force-torque sensor(s). The availability of contact force measurements helps enhance interaction control strategies' performance.

In literature, control strategies are primarily classified into two categories \cite{siciliano2009force}: 1. Direct force control, where contact force control is obtained due to loop closure with force feedback (e.g. pure force control and hybrid force/motion control). 2. Indirect force control, where force control is achieved from motion control, e.g. compliance control (impedance\cite{hogan1985impedance} and admittance control\cite{newman1992stability}). 
In impedance control, a force is controlled after a motion deviation from a set-point is measured, while in admittance, a motion is controlled based on force measurement \cite{keemink2018admittance}. 

The available hardware also offers insights into the selection of the control strategies. Robotic manipulators are either torque-controlled or position-controlled (general term for robots with position and/or velocity control interfaces). Impedance control is possible on torque-controlled robots, whereas admittance control may be implemented on both torque and position-controlled robots. Most industrial manipulators are position-controlled robots. Therefore, admittance control seems to be the critical method for interaction\cite{scherzinger2017forward}. Recently, compliance in such systems has been realized using the forward dynamics compliance control (FDCC) method \cite{scherzinger2021human}, which, owing to its virtual forward dynamics model, can be used with both position and velocity control interfaces. 

In literature, a few authors have emphasized the preference of velocity over position control for robot interaction \cite{duchaine2007general,moreno2003velocity,zelenak2015advantages}. In \cite{duchaine2007general}, position and velocity-based low-level controllers were analyzed on a 3-DOF parallel Cartesian robot. Manipulation of the end-effector by a human operator to experience the virtual dynamics revealed that position-based control is overdamped and virtual mass is much higher than the calibrated mass. Therefore, the velocity controller is better for admittance control. The latter was also found by Zelenak et al. in \cite{zelenak2015advantages}, where Bode plots revealed that at low frequencies, the magnitude of the transfer function of a velocity controller is constant while its phase angle is $\sim$0$^{\circ}$ and, in theory, 
should exhibit smoother compliance in contrast to a position controller. Quantitative analysis was performed on a 1-DOF system to validate the theory \cite{zelenak2015advantages}. However, no direct comparison is provided when extending the concept to multi-DOF serial robots.

On the other hand, several control interaction strategies have been proposed for different applications. Li et al. proposed an admittance controller based on fuzzy adaptive control for force tracking in uncertain environments \cite{li2021fuzzy}. In \cite{ferraguti2019variable} a variable admittance control method is presented. It restores stable behaviour by detecting instabilities and preserving passivity. In \cite{peng2021neural}, Admittance parameters are learnt using reinforcement learning when lacking the knowledge of environmental dynamics. Recently, a novel interaction control paradigm (Forward Dynamics Compliance Control FDCC) based on the idea of forward dynamics simulations\cite{scherzinger2020virtual} on a virtual robotic model is presented in \cite{scherzinger2017forward}. It integrates Admittance, Impedance, and Force control into a combined control architecture. 

 This paper conducts a performance evaluation using the FDCC paradigm to achieve end-effector compliance in a position-controlled robot, when facing the possibility of choosing between position and velocity control interfaces. 
 This paradigm is chosen for its ease of use, simplicity, and robot-agnostic implementation. The evaluation is based on a set of objective \cite{marvel2012best} and subjective experiments involving humans-in-the-loop\cite{duchaine2007general} providing an assessment from a human-robot interaction perspective. Metrics and Artefacts commonly used for benchmarking force control capabilities are used \cite{falco2016benchmarking,patel2015manipulator,falco2015grasping,kimble2020benchmarking}. The main contribution of this paper is to provide a comparative performance analysis of compliance control under the forward dynamics paradigm in position-controlled robots that offer the possibility of choosing between position and velocity hardware control interfaces, such as the widely used UR robot. In the state-of-the-art, very few works have addressed the comparison of position/velocity control interfaces, where velocity control has been preferred for compliance applications. However, no studies have compared compliance control implementations based on different control interfaces for multi-DOF serial manipulators. The latter is crucial because, due to differences in their characteristics, the selection of a specific interface can affect or favour the robot´s response during an interaction.

This paper is organised as follows: Section \ref{sec:Fwd_dynamics} presents the virtual model of FDCC approach and the closed-loop control law. Section \ref{sec:Compliance_control_evaluation} describes the experimental setup and benchmarking metrics. Section \ref{sec:Experiments} presents and discusses the results. Finally, Section \ref{sec:conclusion} offers conclusion and direction of future work.

\section{Virtual Forward Dynamics Based Control} \label{sec:Fwd_dynamics}

\begin{figure}[h!]
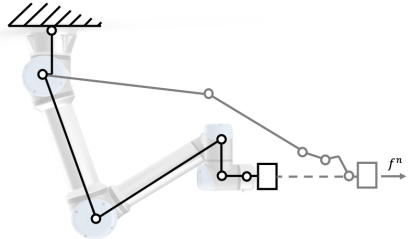

\setlength{\belowcaptionskip}{-15pt}
\centering
       \adjincludegraphics[width=0.69\linewidth,clip]{Figures/FDCC_Sim.png} 
    \caption{Forward Dynamics Illustration. The virtual model (multi-link chain) accelerates in the direction of applied force. Real Robot (ghost) and the Virtual model have the same kinematic structure.}
    \label{fig:virtual_model_representation}
\end{figure}

Forward dynamics models describe how the motion of a body changes due to forces and torques applied to the body. Consider a virtual model of an articulated robotic manipulator as a system of rigid bodies sharing the same kinematic structure as its real counterpart. For this virtual model, forward dynamics computation would describe how the "virtual" robot would move in a "virtual" space under the effect of generalized external forces as illustrated in Fig. \ref{fig:virtual_model_representation}. 
The relationship between these generalized external forces acting on the end-effector $f^n$, torques in the joints $\tau$, and motion in generalized coordinates is described by 
\begin{equation}
    \tau + J^T f^n = H(q)\ddot{q} + C(q,\dot{q}) + G(q)
    \label{eqn:generalized_external_forces}
\end{equation}
where $H$ defines the mechanism's positive definite inertia matrix, $C$ represents the Centrifugal and Coriolis terms, and $G$ denotes the gravitational vector.

\begin{figure}[ht!]
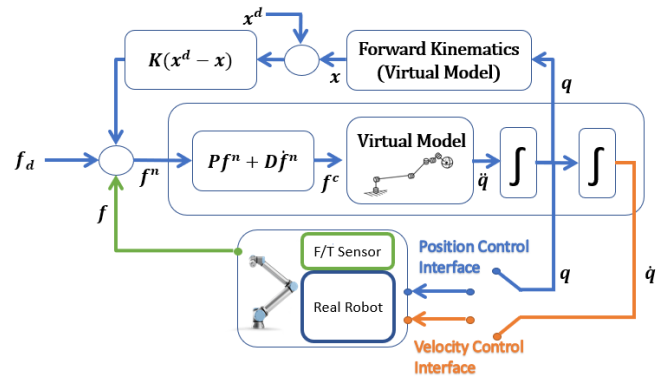

\centering
       \adjincludegraphics[width=\linewidth,clip]{Figures/FDCC_full_BD.png} 
    \caption{Block diagram representation of Forward Dynamics Control of position/velocity controlled manipulators \cite{scherzinger2017forward}}
    \label{fig:FDCC_block}
\end{figure} 

Fig. \ref{fig:FDCC_block} depicts the "virtual model" block for the forward dynamics model. The goal of forward dynamics computation is to solve  \eqref{eqn:generalized_external_forces} for $q(t)$, i.e. to simulate the motion of the model in time under external loads. The behaviour of this virtual model is imposed on a real robot after considering some simplifications based on studying how the controller would perceive the system in closed-loop control. To begin with, for position/velocity-based controllers, gravity is inherently accounted for; therefore, in \eqref{eqn:generalized_external_forces}, the gravity term is dropped\cite{scherzinger2020virtual}. Additionally, it is assumed that the robot accelerates from rest for each control cycle. Therefore, $\dot{q}=0$, and the $C(q,\dot{q})$ can be neglected. Furthermore, the assumption is that external loads are the only loads acting on the model and that these robots do not offer such torque control, so $\tau = 0$. Thus, \eqref{eqn:generalized_external_forces} is reduced to 
 \begin{equation}
    \ddot{q} = H^{-1}(q)J^T f^n
    \label{eqn:simplified_fd_eqn}
\end{equation}

Eq. \eqref{eqn:simplified_fd_eqn} provides a forward mapping from the Cartesian space to the joint space. The virtual model with simplified forward dynamics can now move in response to a net external force. The next step is to formulate a closed-loop control around this model to regulate set-points of forces and motion that describe any interaction state. It involves reference motion tracking in free space in the presence of restoring forces and pure force control (forces guide the robot without any restoring effect). Fig. \ref{fig:FDCC_block} depicts the closed-loop control scheme. Scherzinger uses a PD regulator \cite{scherzinger2021human}:
 
 \begin{equation}
    f^c = P f^n + D \dot{f}^n
    \label{eqn:pd_control_law}
\end{equation}

$P$ and $D$ are positive, semi-definite matrices, $f^n$ the overall net force is now described as follows:

\begin{equation}
    f^n : = f^d - f +K \left( x^d -x \right) + D \left( \dot{x}^d - \dot{x} \right)
        \label{eqn:total_net_force}
\end{equation}

where the $f^d$ is the desired force, $f$ is the force measured by the force/torque sensor, $\dot{x^d}$ and $x^d$ are the desired velocity and the desired position, while as $\dot{x}$ and $x$ are the actual velocity and position information obtained from forward kinematics. The last two terms in  \eqref{eqn:total_net_force} resemble a simplified admittance control law with $K$ and $D$ terms being the stiffness and damping diagonal matrices. Thus, the compliance control law described here is a form of admittance control. Ignoring admittance terms, \eqref{eqn:total_net_force} becomes pure-force control:
\begin{equation}
    f^n : = f^d - f 
        \label{eqn:pure_force}
\end{equation}

In the current implementation of the control framework \cite{CartesianControllers_github}, the direct effect of the damping matrix $D$ is ignored and the net input to the control law in \eqref{eqn:pd_control_law} is:

\begin{equation}
    f^n : = f^d - f +K \left( x^d -x \right)         \label{eqn:total_net_force_wo_damping}
\end{equation}

Thereby imposing a spring-like behaviour on the system. 

To summarize, with \eqref{eqn:total_net_force_wo_damping} as input to the control law described in \eqref{eqn:pd_control_law} the virtual model accelerates. Due to the forward mapping offered by the virtual model, joint accelerations are easily available. These joint accelerations undergo integrations to render motion profiles in the form of joint velocity and joint positions as inputs to the low-level joint controllers of the real robot. These forward dynamics computations are iterative and run completely detached from the real robot's internal motion control cycle.

\section{Performance Evaluation Setup}
\label{sec:Compliance_control_evaluation}

The performance evaluation experiments were performed on a ROS-controlled UR10e robot inside the ZeroG Lab facility at the University of Luxembourg \cite{ZeroG_CVI,muralidharan2022_hitl_iac}. 

\begin{figure}[h!]
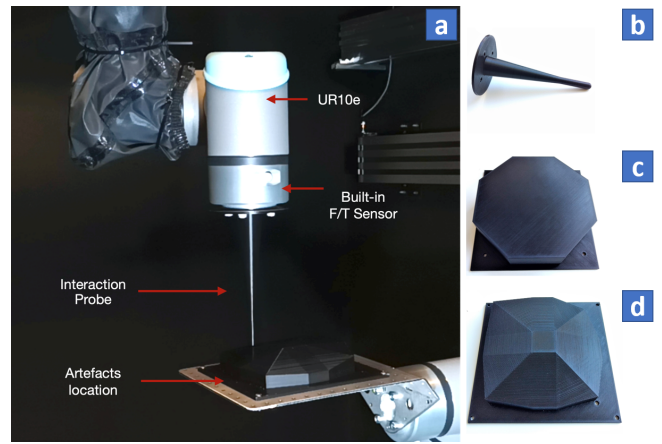

\setlength{\belowcaptionskip}{-10pt}
\centering
       \adjincludegraphics[width=0.99\linewidth,clip]{Figures/UR10eSetup_modified.png} 
    \caption{Experimental Setup. A probe (b) is mounted on a UR10e robot. 3D printed artefacts are used for Disturbance Handling/Obstruction Stability (c) and for Settle Stability (d).}
    \label{fig:exp_scenario}
\end{figure}

The robot was connected over a ROS Network to a control PC installed with Ubuntu 18.04 with a real-time kernel and ROS Melodic to run FDCC algorithms. Each set of experiments was performed for both position and velocity control modes. Force sensor measurements were obtained using the robot's in-built force/torque sensor. During both control modes, the set of parameters depicted in Table \ref{table:parameters} were used to establish a common point of reference for comparison, ensuring that the virtual robot model's dynamics stayed the same for both sets of experiments. This table describes translational and rotational stiffness values per Cartesian axis for compliance parameters in \eqref{eqn:total_net_force_wo_damping}, controller gains for the PD controllers in \eqref{eqn:pd_control_law}, and the choice of values for the Inertia Matrix $H$ of the virtual model \eqref{eqn:simplified_fd_eqn}. Where $m_e$ and $I_e$ are the mass and inertia of the virtual model's last link, while $ml$ and $I_l$ are
the mass and inertia values assigned to the remaining links in the kinematic chain. The kinematic model also included a probe mounted at the robot's flange. Each robot's TCP and payload were correctly calibrated before running the experiments. 
Appropriate wrench transformations were applied to get force/torque measurements in the tool reference frame. Test artefacts for force-control benchmarking, comparable with Falco et al. \cite{falco2016benchmarking}, were used in the experiments. These artefacts were mounted rigidly on a metallic fixture. The fixture was mounted on another available robot to have both desired and accurate positioning of the mounted artefacts. Fig. \ref{fig:exp_scenario} depicts the experimental setup and the 3D printed artefacts.

Metrics and experiments used to characterize the capabilities of FDCC in position and velocity hardware control interfaces are based on  \cite{falco2016benchmarking} and \cite{zelenak2015advantages}. Table \ref{table:Metrics} lists the metrics, their description, and the corresponding test artefacts.

\begin{table*}[h!]
    \caption{Force Control Metrics}
    \label{table:Metrics}
    \centering
    \begin{tabular}{ C{3.5cm}  p{9.3cm}  C{3.5cm} }
         \hline
         \multirow{2}{*}{Metrics} &
         \multirow{2}{*}{\hspace{4cm}Description} & \multirow{2}{*}{Artefact Used} 
         \\ 
          \\
          \hline
          \hline
         Cumulative Work Done (CW) & 
The cumulative work (performed by a human operator or a mechanical device) due to force applied to a compliant robot. Force and the CW is plotted vs. time & $-$ \\
         \hline
         Obstruction Stability (OS) & 
Measurements of the time and force reaction 
associated with an immovable obstruction placed in the path of a robot.
& Disturbance Handling Artefact \\ 
         \hline
         Settle Stability (SS) & 
Calculation of the settling time, overshoot, and steady-state error when
reaching the desired contact force with a surface
& Settle Stability artefact\\ 
         \hline
         Disturbance Handling (DH) & 
The deviation from the desired nominal force is measured when moving
along a surface profile.
& Disturbance Handling Artefact \\ \hline
    \end{tabular}
\end{table*}

\section{Experiments And Results}\label{sec:Experiments}

Experiments were carried out using ROS Implementation of FDCC \cite{CartesianControllers_github} incorporated into the software framework of the ZeroG facility. Data recording was performed using ROSbags and later analysed in Matlab.

\subsection{ Cumulative Work Done (CW)} 

\begin{figure}[h!]
\centering
       \adjincludegraphics[width=0.99\linewidth,trim={0cm 0cm 0cm 0cm},clip]{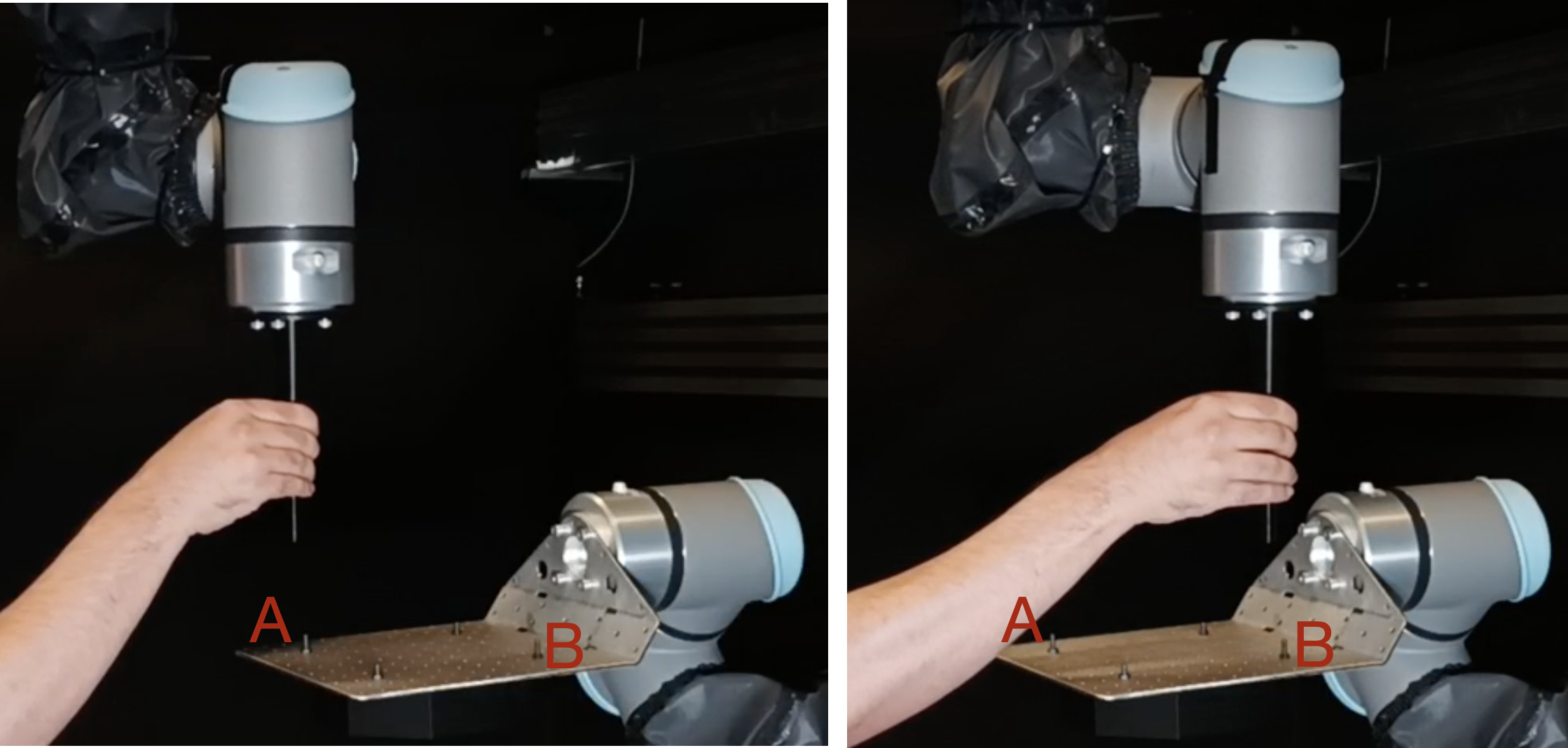} 
    \caption{CW Experiment. Data is gathered when operators move the probe between points A and B.}
    \label{fig:cwd}
\end{figure}

Compliance was enabled on the robot, and four operators were tasked to move the probe tip mounted on the robot back and forth between two markers (A and B in Fig. \ref{fig:cwd}). One back-and-forth motion is classified as one cycle. A total of four cycles were recorded for each of the four trials in the experiment. After the experiment, the operators were asked what they felt about manipulating the robot during velocity-based and position-based control. Besides, forces exerted by the operator were monitored along with the consequent displacement. The latter was to obtain the cumulative work done by the operator.

\begin{figure}[h!]
\centering
       \adjincludegraphics[width=0.99\linewidth,trim={3cm 0cm 3.5cm 1.5cm},clip]{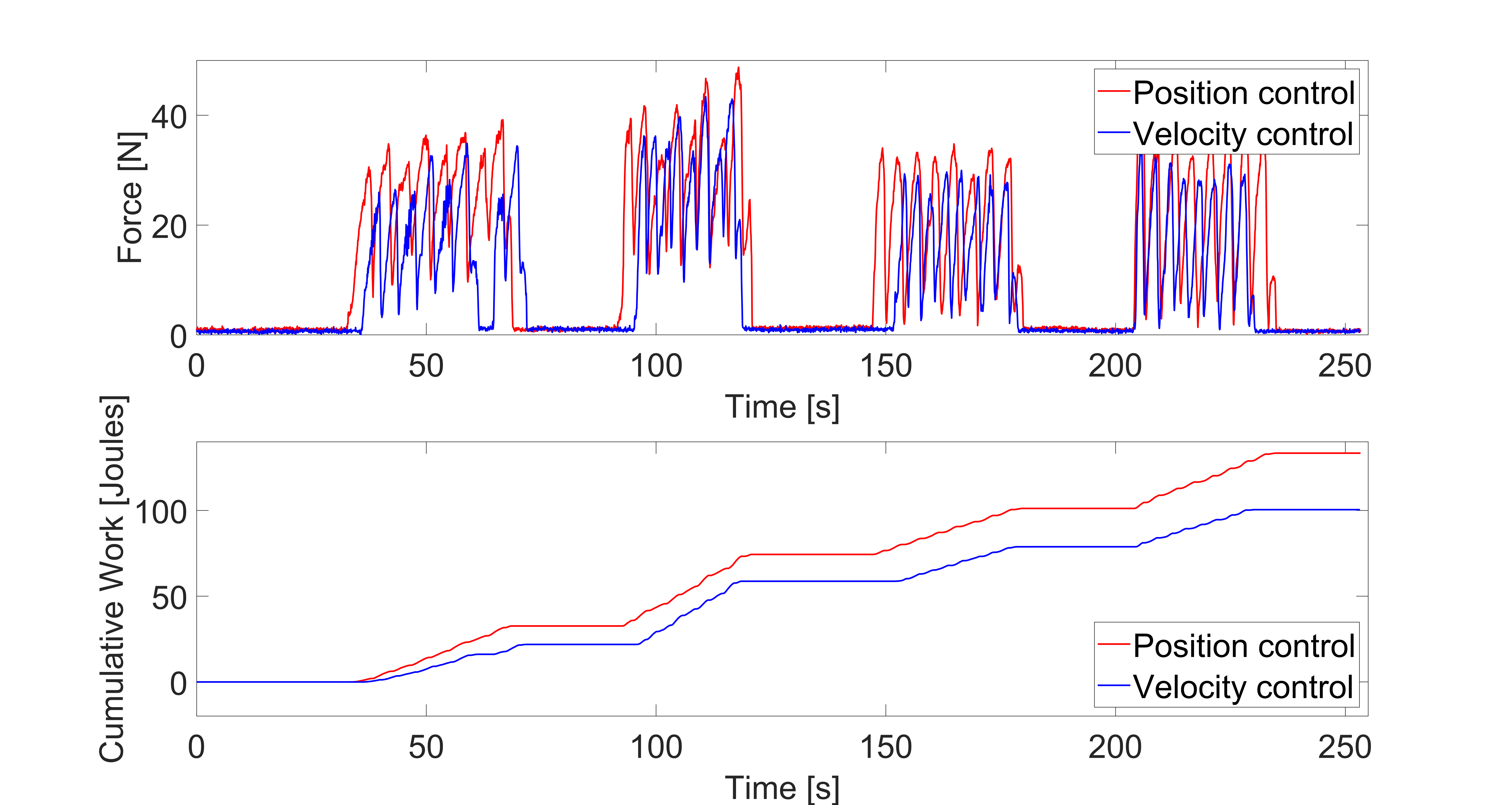} 
    \caption{Force and CW by human operators over time.}
    \label{fig:CW_figure}
\end{figure}

Fig. \ref{fig:CW_figure} shows the results from the trials. The magnitude of force recorded over the four trials indicates that the operator needs to exert less effort when using velocity control than position control.  Consequently, the total work done using velocity control is lower than that of position control. 
Table \ref{table:CW_results} summarizes the reduction in work done per trial due to using velocity control over position control. Besides this, the operators reported that in velocity-control mode the robot felt "more responsive" and less damped than in the  position-control mode. 
The results indicate that for identical compliance parameters of the virtual forward dynamic model, a robot operated in velocity control mode offers an effective reduction in the interaction forces during compliant manipulation in contrast to operating in position control mode. 

\begin{table}[h!]
    \caption{Work Reduction over Trials}
    \label{table:CW_results}
    \centering
    \begin{tabular}{ C{2.0cm} C{4.55cm} }
         \hline
         
         \multirow{2}{*}{Trial} & Work Reduction [Joules]\\ 
         & (position control to velocity control) \\
          \hline
          \hline
         1 &  10.74 (32.93\%)\\ 
         2 &  15.62 (21.01\%)\\ 
         3 &  22.35 (22.10\%)\\ 
         4 &  32.94 (24.70\%)\\
         \rowcolor{gray!30} Average Reduction & 20.41 (25.18\%)\\ \hline 
    \end{tabular}
\end{table}

\subsection{Obstruction Stability (OS)} 

\begin{figure}[h!]
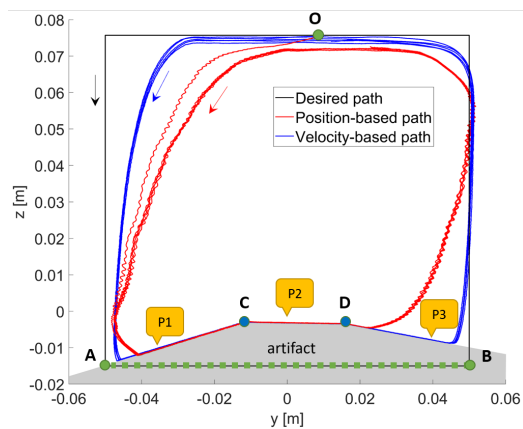

\centering
\setlength{\belowcaptionskip}{-10pt}
       \adjincludegraphics[width=0.8\linewidth,trim={0cm 0cm 0cm 0cm},clip]{Figures/OS/OS_newer.png} 
    \caption{OS Experiments. Comparison (desired vs. actual) of Cartesian paths traversed by the probe tip.}
    \label{fig:OS_actual_path}
\end{figure}

 A set of Cartesian waypoints spaced at equal small intervals, starting at point O and depicted by the black rectangle in Fig. \ref{fig:OS_actual_path}, were imparted as set points to the compliance controller to follow. Four trials were carried out for each control mode in a single run using different parameters (Table \ref{table:parameters}). 
 Fig. \ref{fig:OS_actual_path} illustrates paths the probe tip took during the experiments. 
Owing to the choice of stiffness values depicted in Table \ref{table:parameters}, the probe tip tracked the waypoints with some error. As observed in Fig. \ref{fig:OS_actual_path}, path tracking was less accurate for position control mode.

\begin{table*}[h!]
\caption{FDCC Control Parameters}
\label{table:parameters}
\centering
\begin{tabular}{ p{2.5cm} p{2.0cm}  C{2.1cm} C{2.1cm} C{2.2cm} C{2.2cm}  }
 \hline
 
 & & CW & OS & DH & SS\\
 \hline 
  \hline
 \multirow{2}{2.5cm}{Stiffness} 
  & $\quad$ $K_{x,y,z}$    & 250, 250, 100   & 250, 250, 100    & 1500, 1500, 1500 & 1500, 1500, 1500  \\
  & $\quad$ $K_{Rx,Ry,Rz}$ & 200, 200, 200   & 200, 200, 200    &  200,  200,  200& 200, 200, 200  \\
\hline
 \multirow{6}{2.5cm}{Controller Gains} &$\quad$ $P_x,\; D_x$ & 0.02, 0.0002  & 0.0025, 0.000025 & 0.0025, 0.000025 & 0.0025, 0.000025 \\
 &$\quad$ $P_y,\; D_y$ & 0.02, 0.0002  & 0.0025, 0.000025 & 0.0025, 0.000025 & 0.0025, 0.000025 \\
 &$\quad$ $P_z,\; D_z$ & 0.02, 0.0002  & 0.0025, 0.000025 & 0.0025, 0.000025 & 0.0025, 0.000025 \\
 &$\quad$ $P_{R_x},\; D_{R_x}$ &0.3, 0.002  & 0.035, 0.00025 & 0.035, 0.00025 & 0.035, 0.00025 \\
 &$\quad$ $P_{R_y},\; D_{R_y}$ &0.3, 0.002  & 0.035, 0.00025 & 0.035, 0.00025 & 0.035, 0.00025 \\
 &$\quad$ $P_{R_z},\; D_{R_z}$ &0.3, 0.002  & 0.035, 0.00025 & 0.035, 0.00025 & 0.035, 0.00025\\
\hline
 \multirow{3}{2.5cm}{Inertia Matrix, $H$} 
  & $\quad$ $m_e,\; m_l (kg)$  & $ 1.0, 0.01 $   & $ 1.0, 0.01  $    & $ 1.0 , 0.01  $   & $ 1.0, 0.01  $  \\
  & $\quad$ $I_e (kg m/s^2)$    & $diag[1, 1, 1]$    & $diag[1, 1, 1]$     &   $diag[1, 1, 1]$  & $diag[1, 1, 1]$  \\
  & $\quad$ $I_l (kg m/s^2)$    & $10^{-6}I_e $      & $10^{-6}I_e $       & $10^{-6}I_e $      & $10^{-6}I_e  $\\
  
  
\hline
\end{tabular}
\end{table*}

The path of interest was the linear segment AB passing through the artefact (green dots). Starting from point A, the probe tip slid over the artefact deviating from the desired path and grazing over the artefact. In both modes, the probe tip compliantly followed the obstacle profile along the three angled surfaces P1, P2 and P3, as shown in Fig. \ref{fig:OS_actual_path}.

\begin{figure}[h!]

     \centering
     \begin{subfigure}[b]{1.0\linewidth}
         \centering
         \adjincludegraphics[width=\linewidth,trim={0cm 0cm 0cm 0cm},clip]{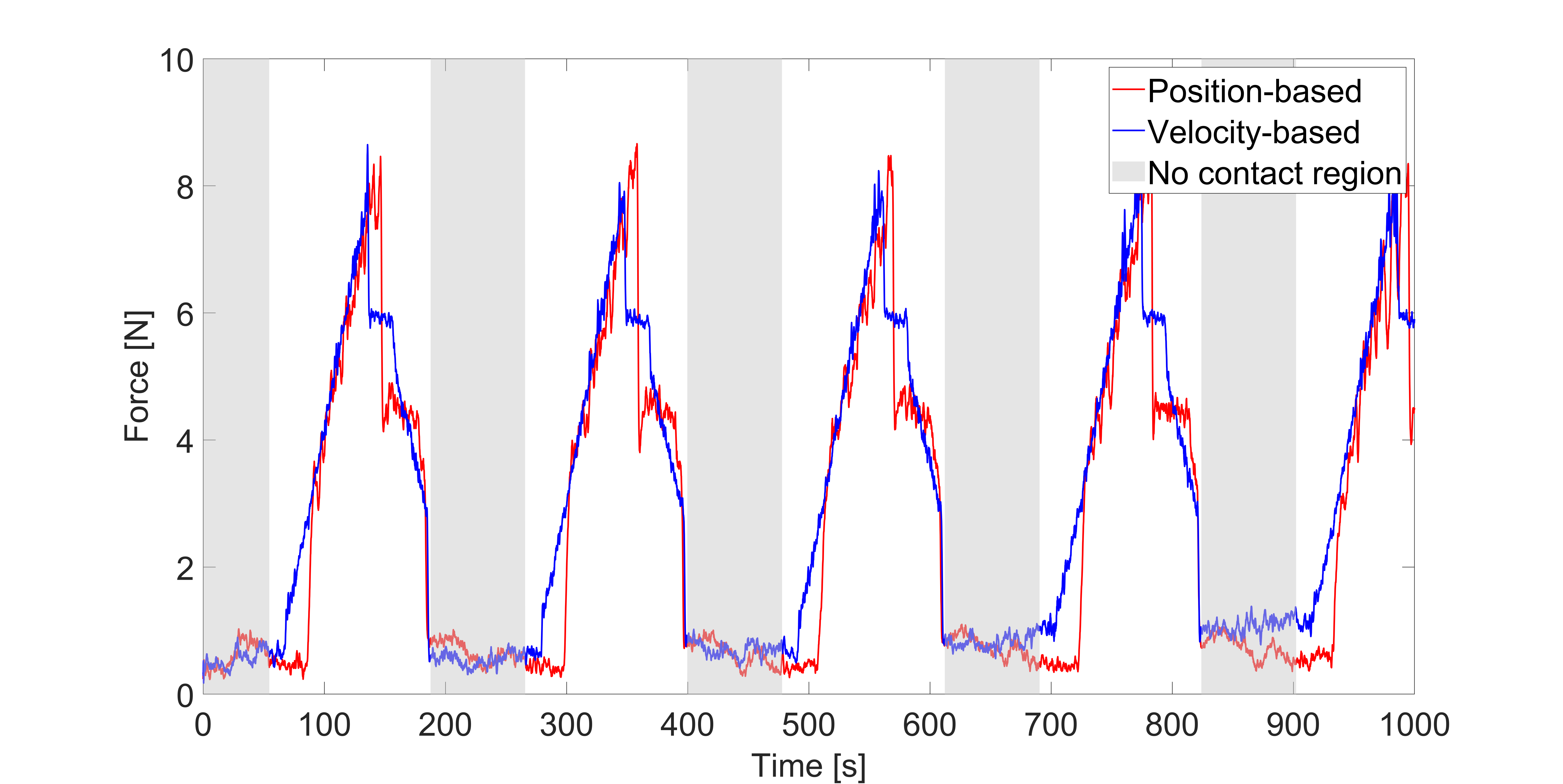}
         \caption{Force history over all trials}
         \label{fig:OS_force_history}
     \end{subfigure}
     \\
     \centering
     \begin{subfigure}[b]{1.0\linewidth}
         \centering
         \adjincludegraphics[width=\linewidth,trim={0cm 0cm 0cm 0cm},clip]{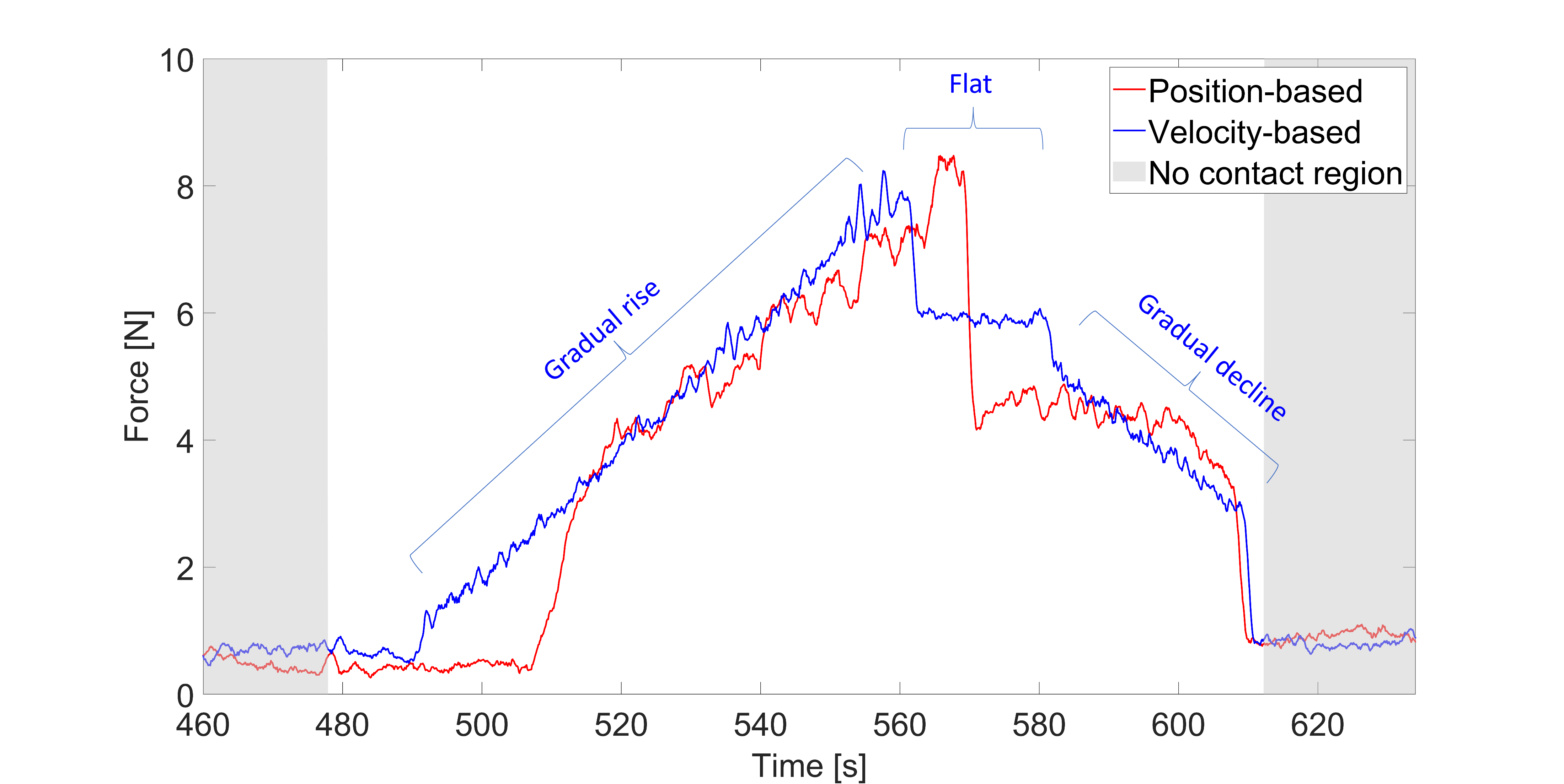}
         \caption{Force history (zoomed-in view) as the tool tip progresses over distinct regions along the artefact}
         \label{fig:OS_force_zoom}
     \end{subfigure}
     \caption{Force profile for Obstruction Stability experiments}
     \label{fig:OS_force}
\end{figure}

Table \ref{table:os} presents the results.
Fig. \ref{fig:OS_force_zoom} depicts the magnitude of the force profile during one of the trials as the probe tip moves over the artefact's angled surfaces, trying to follow the segment AB. Although contact was consistently maintained in all experiments between the probe and the artefact, the results revealed more subtle details about the contact. In velocity control mode, the probe tip maintained a consistent, stable contact as the force rose steadily along the surface P1, maintained almost a constant magnitude along P2, and gradually decreased along the surface P3. During position control mode, the force profile has more variation during the rise along P1. A higher force is incurred at the transition to P2, followed by a drastic decrease. The force then maintained a constant value with some variation, followed by a gradual decrease, which was cut short due to lifting the probe off the surface in an attempt to track the desired path. 

\begin{table}[h!]
\caption{Obstruction Stability Results}
\label{table:os} 
\centering
\begin{tabular}{ C{1.4cm} C{1.4cm} C{1.4cm} C{1.4cm}  C{1.2cm} }
 \hline
 Mode &  Smooth Contact & Lower Incurred Force & Time to reach goal (s)  \\
 \hline 
  \hline
 \multirow{2}{1.0cm}{Position Control} 
  & \multirow{2}{1.0cm}{$\quad$ $-$}    &  \multirow{2}{1.0cm}{$\quad$ $-$}  &   \multirow{2}{1.0cm}{$\quad$ $30.0$}   \\ 
  &  &  &    \\

\hline
 \multirow{2}{1.0cm}{Velocity Control} 
  & \multirow{2}{1.0cm}{$\quad$ $\checkmark$}    & \multirow{2}{1.0cm}{$\quad$ $\checkmark$}    & \multirow{2}{1.0cm}{$\quad$ $20.0$}     \\ 
  & &   &    \\

\hline
\end{tabular}
\end{table}

As the probe did not go through position A or B precisely, the time taken to reach the goal could not be established using these points. Instead, points C and D were chosen to measure the time the probe moved between C and D to infer in which mode the goal would be reached in a shorter time. In velocity control mode, it took 20 seconds, while in position control, it took 30 seconds. Based on this, it is inferred that a goal can be reached faster when operating under velocity control in the presence of an obstacle.

\subsection{Settle Stability (SS)} 

\begin{figure}[h!]
\centering

       \adjincludegraphics[width=0.79\linewidth,trim={3cm 0cm 3.5cm 0 cm},clip]{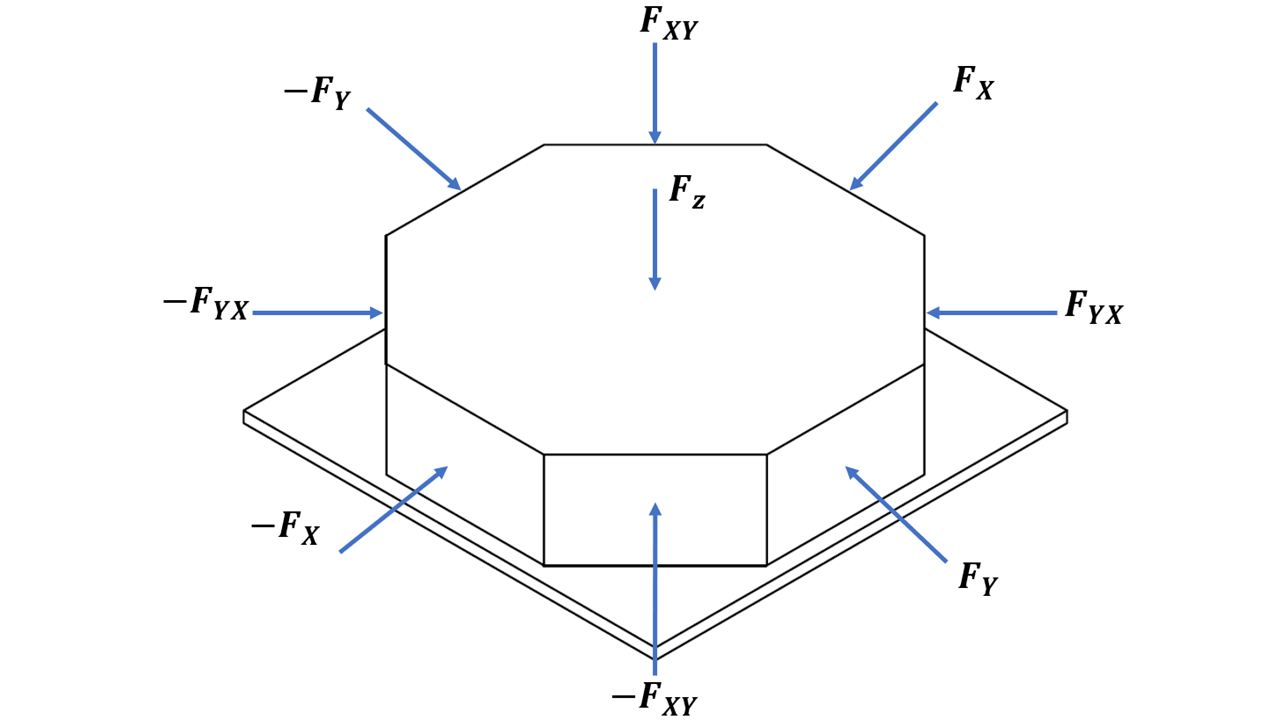} 
    \caption{Force directions applied on the SS artefact.}
    \label{fig:SS_ss}
\end{figure}

The probe tip was commanded with a step input force of various magnitudes to observe the settle stability metrics. Three trials each were carried out applying step input forces in  \textit{x},  \textit{y} and  \textit{z} axes as well as in the  \textit{xy}-plane of the probe tool tip (as shown in Fig. \ref{fig:SS_ss}). For each trial, the desired force level was expected to be maintained while making contact with the surface. The test began from a non-contact state. The probe started a few millimetres away from the surface. On applying force, the probe would tap the surface of interest. 

The contact forces were recorded for the desired input force levels described in the Table \ref{table:ss} (see the column for Step Force) depending upon 1) minimum force applied to evoke a useful robot response, 2) Maximum applied force depending on the robot's payload capability and stiffness of the probe tip. 3) Mid values between the minimum and the maximum applied forces. The maximum applied force in the \textit{x}, \textit{y} and \textit{z} axis was chosen as 40 N, 40 N and 70 N, respectively, to avoid breakage of the probe tip during the experiments. Moreover, on the z-axis, two mid-range values (25 N and 40 N) were chosen instead of one due to a more extensive range of applied possible forces along this axis.

\begin{figure}[h!]

    \centering
     \begin{subfigure}[b]{0.90\linewidth}
         \centering
         \adjincludegraphics[width=\linewidth,trim={0cm 0cm 0cm 0cm},clip]{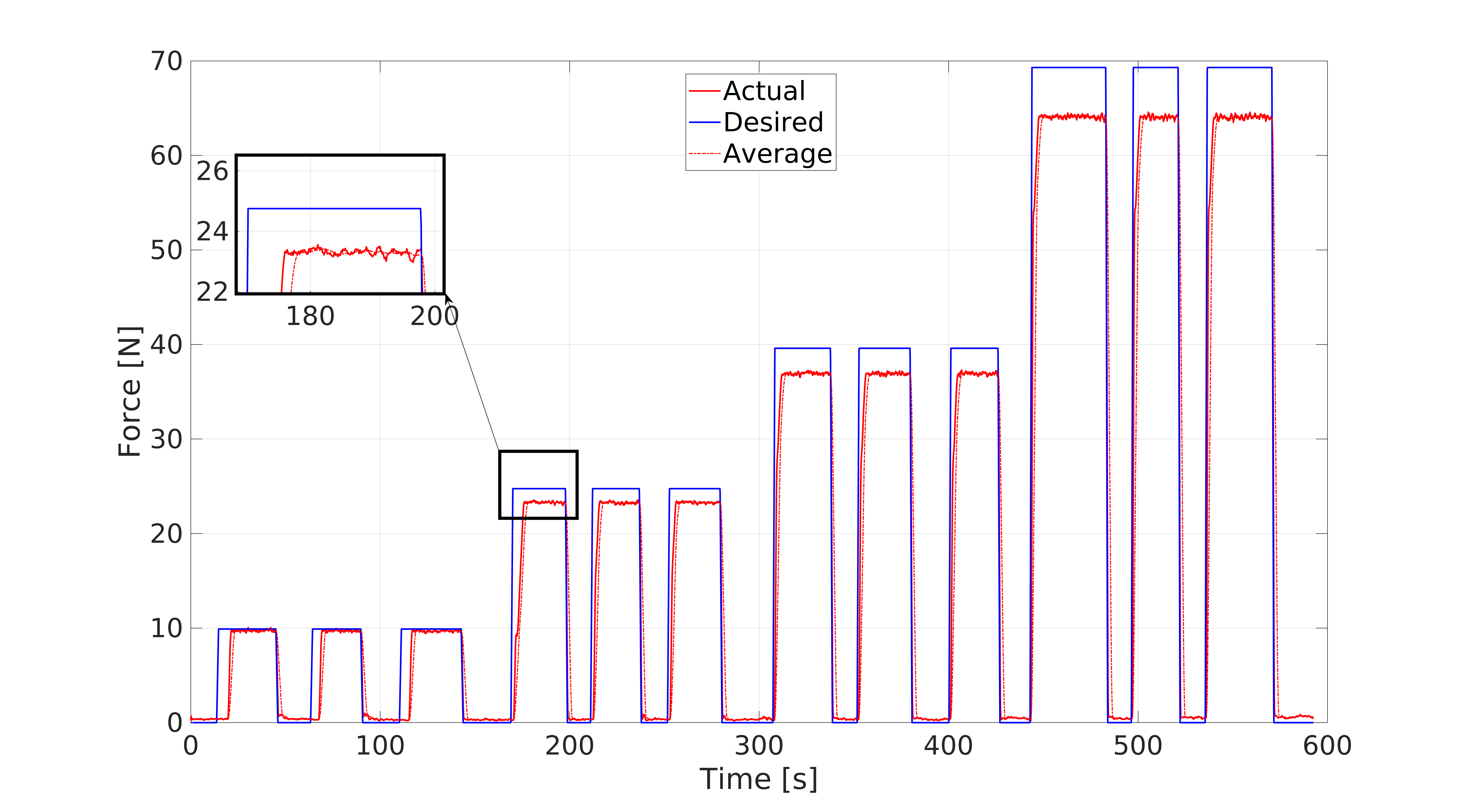}
         \caption{Velocity controlled}
         \label{fig:SS_force_response_vel}
     \end{subfigure}
     \\
     \centering
     \begin{subfigure}[b]{0.90\linewidth}
         \centering
         \adjincludegraphics[width=\linewidth,trim={0cm 0cm 0cm 0cm},clip]{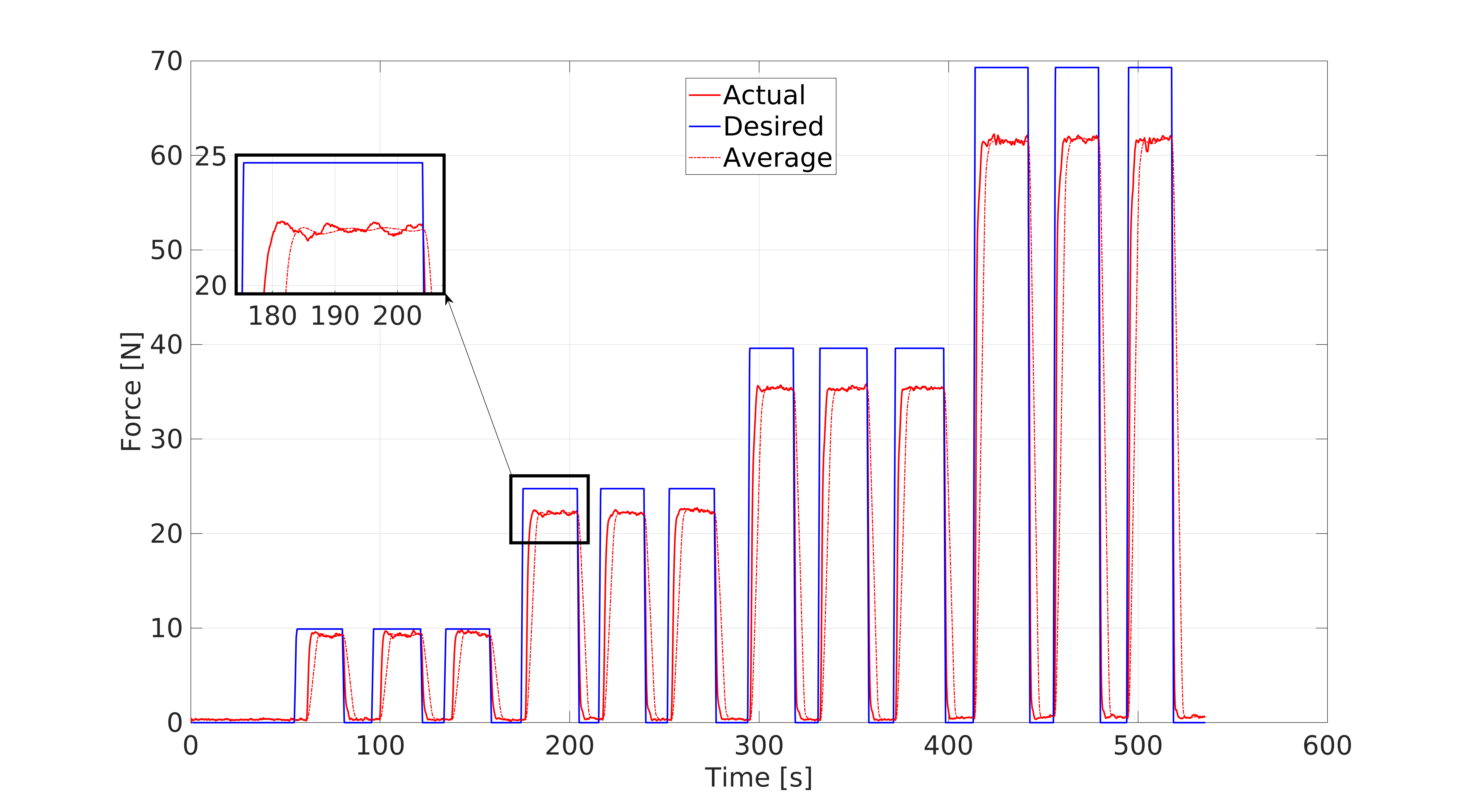}
         \caption{Position controlled}
         \label{fig:SS_force_response_pos}
     \end{subfigure}
    \caption{Force Response for Settle Stability experiments}
    \label{fig:SS_force_response}
\end{figure}

Table \ref{table:ss} reports the overshoot, settling time and steady-state performance of the same control algorithm under velocity and position control interfaces. Fig. \ref{fig:SS_force_response} illustrates the force response plot for both position and velocity control mode when step input forces of 10, 25, 40 and 70 N were applied to the artefact along z-axis (Corresponding results are reported in the shaded rows in Table \ref{table:ss}).

\begin{table*}[h!]
\fontsize{11}{9}
\caption{Settle Stability}
\label{table:ss}
\centering
\begin{tabular}[c]{C{1.5cm}  C{1.1cm}  C{1.8cm}  R{1.0cm}  R{1.0cm}  R{1.0cm}  R{1.0cm}  R{1.0cm}  R{1.0cm} }
\hline 
\multirow{2}{*}{Mode} &
  \multirow{2}{*}{Direction} &
  \multirow{2}{*}{Step Force [N]} &
  \multicolumn{2}{c }{Overshoot [N]} &
  \multicolumn{2}{c }{Settling time [s]} &
  \multicolumn{2}{c }{Steady state error [N]} \\ \cline{4-5} 
  \cline{6-7}
   \cline{8-9}
                           &    &    & Mean & SD   & Mean & SD   & Mean   & SD   \\ \hline 
                           \hline 
 \multirow{13}{*}{\makecell{Position \\Control} } & X  & 10 & 0.54 & 0.17 & 7.00 & 0.60 & 0.74  & 0.05 \\
                           & X  & 25 & 1.26 & 0.15 & 6.83 & 0.45 & 4.03  & 0.09 \\
                           & X  & 40 & 1.12 & 0.12 & 7.00 & 0.17 & 7.17  & 0.03 \\ \cline{2-9}
                           & Y  & 10 & 0.52 & 0.20 & 8.10 & 0.66 & 5.90  & 0.05 \\
                           & Y  & 25 & 1.39 & 0.10 & 8.17 & 1.00 & 9.95  & 0.06 \\
                           & Y  & 40 & 1.30 & 0.07 & 8.43 & 0.59 & \cellcolor{cyan!30}13.33 & 0.10 \\ \cline{2-9} 
                           & XY & 10 & 1.06 & 0.17 & 7.07 & 0.21 & 4.46  & 0.13 \\
                           & XY & 25 & 1.22 & 0.06 & 7.10 & 0.10 & 3.15  & 0.14 \\
                           & XY & 40 & 1.35 & 0.23 & 7.23 & 0.12 & 4.62  & 0.08 \\ \cline{2-9}
                           \rowcolor{gray!30} \cellcolor{white} & Z  & 10 & \cellcolor{red!25}0.36 & 0.07 & \cellcolor{green!25}6.47 & 0.15 & \cellcolor{blue!25}1.68  & \cellcolor{cyan!30}0.37 \\
                           \rowcolor{gray!30} \cellcolor{white} & Z  & 25 & \cellcolor{red!25}0.35 & 0.05 & \cellcolor{green!25}8.20 & 0.53 & \cellcolor{blue!25}2.67  & 0.14 \\
                           \rowcolor{gray!30} \cellcolor{white} & Z  & 40 & \cellcolor{red!25}0.17 & 0.14 & \cellcolor{green!25}8.23 & 0.23 & \cellcolor{blue!25}4.37  & 0.03 \\
                           \rowcolor{gray!30} \cellcolor{white} & Z  & 70 & \cellcolor{red!25}0.40 & 0.28 & \cellcolor{green!25}8.60 & 0.20 & \cellcolor{blue!25}8.12  & 0.12 \\ 
                           \hline 
                           
\multirow{13}{*}{\makecell{Velocity \\Control} } & X  & 10 & 0.16 & 0.03 & 3.43 & 0.06 & 0.77   & 0.05 \\
                           & X  & 25 & 0.17 & 0.06 & 3.93 & 0.15 & 2.50  & 0.13 \\
                           & X  & 40 & 0.26 & 0.08 & 4.20 & 0.10 & 5.41  & 0.06 \\ \cline{2-9} 
                           & Y  & 10 & 0.13 & 0.07 & 4.00 & 0.53 & 6.09  & 0.02 \\
                           & Y  & 25 & 0.16 & 0.05 & 3.60 & 0.10 & 8.67  & 0.05 \\
                           & Y  & 40 & 0.17 & 0.03 & 3.80 & 0.10 & \cellcolor{cyan!30}11.22 & 0.02 \\ \cline{2-9} 
                           & XY & 10 & 0.15 & 0.09 & 3.73 & 0.23 & 6.03  & 0.02 \\
                           & XY & 25 & 0.19 & 0.08 & 3.87 & 0.06 & 3.85  & 0.03 \\
                           & XY & 40 & 0.29 & 0.04 & 3.93 & 0.23 & 4.37  & \cellcolor{cyan!30}0.52 \\ \cline{2-9}
                           \rowcolor{gray!30} \cellcolor{white} & Z  & 10 & \cellcolor{red!25}0.25 & 0.12 & \cellcolor{green!25}3.77 & 0.12 & \cellcolor{blue!25}0.89  & 0.07 \\
                          \rowcolor{gray!30} \cellcolor{white} & Z  & 25 & \cellcolor{red!25}0.17 & 0.06 & \cellcolor{green!25}5.50 & 0.10 & \cellcolor{blue!25}1.80  & 0.03 \\
                           \rowcolor{gray!30} \cellcolor{white} & Z  & 40 & \cellcolor{red!25}0.18 & 0.19 & \cellcolor{green!25}\cellcolor{green!25}5.87 & 0.25 & \cellcolor{blue!25}2.98  & 0.02 \\
                           \rowcolor{gray!30} \cellcolor{white} & Z  & 70 & \cellcolor{red!25}0.23 & 0.13 & \cellcolor{green!25}6.00 & 0.17 & \cellcolor{blue!25}5.54  & 0.05 \\ \hline 
\end{tabular}
\end{table*}

Overshoot force value is determined from the difference between the maximum force value detected prior to the steady state force value.

Comparing Figs. \ref{fig:SS_force_response_vel} and \ref{fig:SS_force_response_pos} the overshoot in position control is much more pronounced than in velocity control. The highest peak overshoots for position control was observed for \textit{x}, \textit{y} and \textit{xy}-planes although it stayed under 2 N. The lowest overshoots were observed along z-axis. Overshoots for velocity control stayed under 1 N. These low overshoots can be attributed to the choice of stiffness parameters and PD Controller gains of the virtual forward dynamics controller. 

Settling time is calculated as the time between the probe's impact on the obstructing artefact once the step force input is applied and the time taken to reach a steady state.

Settling times were consistent throughout the experiments. However, velocity-based control yielded lower settling times than position-based control. The shortest settling time (6.47 and 3.43 seconds for position and velocity control, respectively) were reported along their corresponding \textit{z} and \textit{x} axes. The longest settling times (8.6 and 6.0 seconds) were reported for both cases along the z-axis.   

The steady-state error was calculated as the RMSE between the desired force input and the actual measured force. 

At higher step forces, the steady-state errors increased in magnitude. The latter was a common observation for both position and velocity control. Force tracking was accurate for position control within 13 N from the desired force. For velocity control, the accuracy was within 11 N. The experiments demonstrated repeatability with a standard deviation lower or close to 0.5 N. The highest steady-state error was reported along the y-axis for both modes (13.33 N for position control and 11.22 for velocity). These errors, in general, were also higher in position control than in velocity control.

\subsection{Disturbance Handling (DH)} 
\begin{figure}[h!]
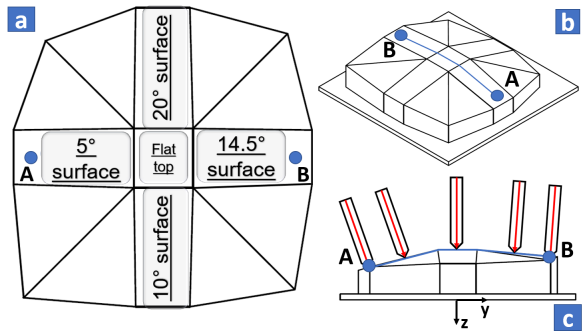

\centering
\setlength{\belowcaptionskip}{-10pt}
       \adjincludegraphics[width=0.89\linewidth,trim={0cm 0cm 0cm 0cm},clip]{Figures/DH/full_dh_pic.png}
    \caption{Disturbance Handling Artefact: (a)Angled surfaces (b) Path of Interest AB (c) Probe's constant contact normal to the surface throughout the motion. }
    \label{fig:DH_dh_4}
\end{figure}

These experiments were performed using the disturbance handling artefact having angled planar surfaces as seen in Fig.  \ref{fig:DH_dh_4}. The probe tip was made to have constant contact with the surface, maintaining a force applied normal to the artefact's face. The tip then moved along a Cartesian trajectory along a chosen segment along the artefact profile (with A as starting point and B as the end; depicted in Figs. \ref{fig:DH_dh_4} b and c).

\begin{figure}[h!]
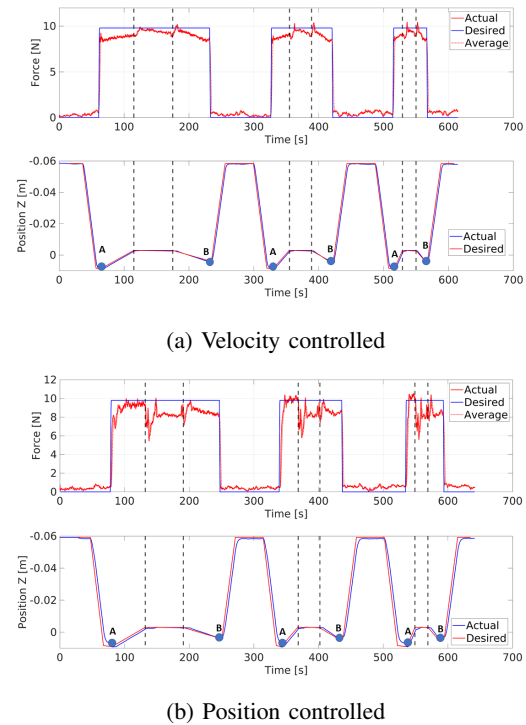

     \centering
     \begin{subfigure}[b]{0.90\linewidth}
         \centering
         \adjincludegraphics[width=\linewidth,trim={0cm 0cm 0cm 0cm},clip]{Figures/DH/DH_Results/force_motion_comparison_VB.png}
         \caption{Velocity controlled}
         \label{fig:DH_force_motion_VB}
     \end{subfigure}
     \\
     \centering
     \begin{subfigure}[b]{0.90\linewidth}
         \centering
         \adjincludegraphics[width=\linewidth,trim={0cm 0cm 0cm 0cm},clip]{Figures/DH/DH_Results/force_motion_comparison_PB.png}
         \caption{Position controlled}
         \label{fig:DH_force_motion_PB}
     \end{subfigure}
     \caption{Force response and Motion Profiles during Disturbance Handling Experiments.}
     \label{fig:DH_force_motion}
\end{figure}

The robot was commanded to apply the desired force of 10 N for each control interface. These tests were performed at three different operational speeds. The results comparing the metrics are presented in Fig. \ref{fig:DH_force_motion} and Table  \ref{table:dh}. Fig. \ref{fig:DH_force_motion} presents the probe tip's force profile and motion profile for z-axis. The vertical dotted line marks the transition between the angled surfaces. The force profile for velocity control is smoother than that of position control. In general, an increase in the operational speed caused more jerky interaction, although contact was maintained. The contact forces varied mainly during the transition from one angled surface to the next and during the descending motion.

In general, the RMSE and Controller RMSE values for both control modes increased with the speed of the probe tip. In position control, the Controller's force tracking performance was within 2 N of desired force, while in the case of velocity control, it was within 1 N of the desired force. The proximity between the Controller RMSE and the Total RMSE tells about the source of the error. Closer the Controller RMSE is to the total RMSE, the error is more due to the Controller than it is due to the force feedback. For velocity control, the error due to force feedback seemed more pronounced than position control. Overall, better force-tracking and disturbance handling is observed for velocity control over position control. 

\begin{table}[h!]
\caption{Disturbance Handling}
\label{table:dh} 
\centering
\begin{tabular}{ C{1.4cm} C{1.2cm}  C{1.2cm} C{1.2cm}  C{1.4cm} }
 \hline
 Mode & Step Input Force (N) & Speed (mm/s) & Total RMSE (N) & Controller RMSE (N) \\
 \hline 
  \hline
 \multirow{3}{1.0cm}{Position Control} 
  & \multirow{3}{1.0cm}{$\quad$ $10$}    & 0.6  &  1.608    & 1.143   \\ 
  &  & 1.0 & 1.925  &  1.629 \\  
  &  & 2.0 & 2.000  &  1.758\\ 
  
\hline
 \multirow{3}{1.0cm}{Velocity Control} 
  & \multirow{3}{1.0cm}{$\quad$ $10$}    &  0.6  &  0.835   & 1.147    \\ 
  & & 1.0  & 0.898 & 1.115  \\  
  & & 2.0 & 0.903 & 1.534 \\

\hline
\end{tabular}
\end{table}

\subsection{Discussion}
The cumulative work experiment demonstrated that FDCC implemented over velocity control interface is more efficient in terms of total work done, smoothness of motion, and reduction of interaction forces. The results for cumulative work done are in agreement with similar work done in literature. Obstruction Stability experiments proved that FDCC over velocity control maintained smoother contact, incurred lower forces, and reached the goal efficiently when in presence of an immovable obstacle. Steady-stability experiments established stability of velocity based FDCC control over position based control owing to the former's lower overshoot, settling times and steady-state errors. Finally, disturbance handling experiments established the superior capability of FDCC to perform better force tracking and disturbance rejection over position control based FDCC. Given the possible payload capacity and the operational speeds of the robot, it would have been possible to perform these experiments at higher desired forces and operational speeds. However, the probe material did not permit to explore full capabilities, as it exhibited, albeit, low but observable compliance laterally.

\section{Conclusion}
\label{sec:conclusion}

This paper presented a performance evaluation of FDCC considering the choice of control interface at the hardware level in a position-controlled robot. Experiments were performed using identical compliance control parameters, each time operating the robot either with velocity or position control interfaces. Metrics commonly used for benchmarking force control capabilities were used to assess the controllers. Results showed that for end-effector compliance in position-controlled robots, velocity control interface and hence velocity-based compliance exhibit better performance than position-based compliance. Our future work will focus on testing the choice of interfaces and implementation of FDCC approach for on-orbit assembly tasks.

\bibliographystyle{ieeetr}
\bibliography{references}

\begin{thebibliography}{10}

\bibitem{siciliano2009force}
B.~Siciliano, L.~Sciavicco, L.~Villani, and G.~Oriolo, {\em Force control}.
\newblock Springer, 2009.

\bibitem{papadopoulos2021robotic}
E.~Papadopoulos, F.~Aghili, O.~Ma, and R.~Lampariello, ``Robotic manipulation
  and capture in space: A survey,'' {\em Frontiers in Robotics and AI}, p.~228,
  2021.

\bibitem{buckmaster2008compliant}
D.~J. Buckmaster, W.~S. Newman, and S.~D. Somes, ``Compliant motion control for
  robust robotic surface finishing,'' in {\em 2008 7th World Congress on
  Intelligent Control and Automation}, pp.~559--564, IEEE, 2008.

\bibitem{gitai_space_taskboard}
``Robotic hand built by japanese company to be part of tech demo on space
  station in 2021.'' Available at
  \url{https://hitechglitz.com/robotic-hand-built-by-japanese-company-to-be-part-of-tech-demo-on-space-station-in-2021-technology-news-firstpost/
  }.
\newblock Accessed: September 14, 2022.

\bibitem{nist_taskboard}
``Nist assembly task board.'' Available at
  \url{https://www.uml.edu/research/nerve/nist-assembly-task-board-form.aspx}.
\newblock Accessed: September 14, 2022.

\bibitem{surface_finshing}
``Surface finishing kit.'' Available at
  \url{https://robotiq.com/products/surface-finishing-kit}.
\newblock Accessed: September 14, 2022.

\bibitem{hogan1985impedance}
N.~Hogan, ``{Impedance Control: An Approach to Manipulation: Part
  II—Implementation},'' {\em Journal of Dynamic Systems, Measurement, and
  Control}, vol.~107, pp.~8--16, 03 1985.

\bibitem{newman1992stability}
W.~S. Newman, ``{Stability and Performance Limits of Interaction
  Controllers},'' {\em Journal of Dynamic Systems, Measurement, and Control},
  vol.~114, pp.~563--570, 12 1992.

\bibitem{keemink2018admittance}
A.~Q. Keemink, H.~van~der Kooij, and A.~H. Stienen, ``Admittance control for
  physical human--robot interaction,'' {\em The International Journal of
  Robotics Research}, vol.~37, no.~11, pp.~1421--1444, 2018.

\bibitem{scherzinger2017forward}
S.~Scherzinger, A.~Roennau, and R.~Dillmann, ``Forward dynamics compliance
  control (fdcc): A new approach to cartesian compliance for robotic
  manipulators,'' in {\em 2017 IEEE/RSJ International Conference on Intelligent
  Robots and Systems (IROS)}, pp.~4568--4575, IEEE, 2017.

\bibitem{scherzinger2021human}
S.~Scherzinger, {\em Human-Inspired Compliant Controllers for Robotic
  Assembly}.
\newblock PhD thesis, Karlsruher Institut f{\"u}r Technologie (KIT), 2021.

\bibitem{duchaine2007general}
V.~Duchaine and C.~M. Gosselin, ``General model of human-robot cooperation
  using a novel velocity based variable impedance control,'' in {\em Second
  Joint EuroHaptics Conference and Symposium on Haptic Interfaces for Virtual
  Environment and Teleoperator Systems (WHC'07)}, pp.~446--451, IEEE, 2007.

\bibitem{moreno2003velocity}
J.~Moreno and R.~Kelly, ``Velocity control of robot manipulators: analysis and
  experiments,'' {\em International Journal of Control}, vol.~76, no.~14,
  pp.~1420--1427, 2003.

\bibitem{zelenak2015advantages}
A.~Zelenak, C.~Peterson, J.~Thompson, and M.~Pryor, ``The advantages of
  velocity control for reactive robot motion,'' in {\em Dynamic Systems and
  Control Conference}, vol.~57267, American Society of Mechanical Engineers,
  2015.

\bibitem{li2021fuzzy}
Z.~Li, H.~Huang, X.~Song, W.~Xu, and B.~Li, ``A fuzzy adaptive admittance
  controller for force tracking in an uncertain contact environment,'' {\em IET
  Control Theory \& Applications}, vol.~15, no.~17, pp.~2158--2170, 2021.

\bibitem{ferraguti2019variable}
F.~Ferraguti, C.~Talignani~Landi, L.~Sabattini, M.~Bonf{\`e}, C.~Fantuzzi, and
  C.~Secchi, ``A variable admittance control strategy for stable physical
  human--robot interaction,'' {\em The International Journal of Robotics
  Research}, vol.~38, no.~6, pp.~747--765, 2019.

\bibitem{peng2021neural}
G.~Peng, C.~P. Chen, and C.~Yang, ``Neural networks enhanced optimal admittance
  control of robot-environment interaction using reinforcement learning,'' {\em
  IEEE Transactions on Neural Networks and Learning Systems}, 2021.

\bibitem{scherzinger2020virtual}
S.~Scherzinger, A.~Roennau, and R.~Dillmann, ``Virtual forward dynamics models
  for cartesian robot control,'' {\em arXiv preprint arXiv:2009.11888}, 2020.

\bibitem{marvel2012best}
J.~Marvel, J.~Falco, and J.~Marvel, {\em Best practices and performance metrics
  using force control for robotic assembly}.
\newblock Citeseer, 2012.

\bibitem{falco2016benchmarking}
J.~Falco, J.~Marvel, R.~Norcross, K.~Van~Wyk, {\em et~al.}, ``Benchmarking
  robot force control capabilities: Experimental results,'' {\em National
  Institute of Standards and Technology (NIST)}, vol.~100, 2016.

\bibitem{patel2015manipulator}
S.~Patel and T.~Sobh, ``Manipulator performance measures-a comprehensive
  literature survey,'' {\em Journal of Intelligent \& Robotic Systems},
  vol.~77, no.~3, pp.~547--570, 2015.

\bibitem{falco2015grasping}
J.~Falco, K.~Van~Wyk, S.~Liu, and S.~Carpin, ``Grasping the performance:
  Facilitating replicable performance measures via benchmarking and
  standardized methodologies,'' {\em IEEE Robotics \& Automation Magazine},
  vol.~22, no.~4, pp.~125--136, 2015.

\bibitem{kimble2020benchmarking}
K.~Kimble, K.~Van~Wyk, J.~Falco, E.~Messina, Y.~Sun, M.~Shibata, W.~Uemura, and
  Y.~Yokokohji, ``Benchmarking protocols for evaluating small parts robotic
  assembly systems,'' {\em IEEE robotics and automation letters}, vol.~5,
  no.~2, pp.~883--889, 2020.

\bibitem{CartesianControllers_github}
``Cartesian controllers.''
  \url{https://github.com/fzi-forschungszentrum-informatik/cartesian_controllers}.
\newblock Accessed: August 30, 2022.

\bibitem{ZeroG_CVI}
L.~Pauly, M.~L. Jamrozik, M.~O. Del~Castillo, O.~Borgue, I.~P. Singh, M.~R.
  Makhdoomi, O.-O. Christidi-Loumpasefski, V.~Gaudilliere, C.~Martinez,
  A.~Rathinam, {\em et~al.}, ``Lessons from a space lab--an image acquisition
  perspective,'' {\em arXiv preprint arXiv:2208.08865}, 2022.

\bibitem{muralidharan2022_hitl_iac}
V.~Muralidharan, M.~R. Makhdoomi, K.~R. Barad, L.~M. Amaya-Mej\'{i}a, K.~C.
  Howell, C.~Martinez~Luna, and M.~A. Olivares~Mendez, ``{Hardware-in-the-loop
  Proximity Operations in Cislunar Space},'' in {\em International
  Astronautical Congress (IAC), Paris, France}, 2022.

\end{thebibliography}

\vspace{12pt}
\end{document}